\documentclass[runningheads]{llncs}
\usepackage{graphicx}

\usepackage{dirtytalk}
\usepackage{amsmath}
\usepackage{url}
\usepackage{hyperref}

\usepackage{tikz}

\usepackage{color}
\usepackage{lingmacros}
\begin{document}
\title{Toponym Identification in Epidemiology Articles -- A~Deep Learning Approach}
%
%
\author{MohammadReza Davari\and Leila Kosseim\and Tien D. Bui}
%
%
\institute{Dept. Computer Science and Software Engineering \\Concordia University, Montreal QC H3G 1M8, Canada \email{mohammadreza.davari@mail.concordia.ca}\\\email{\{leila.kosseim,tien.bui\}@concordia.ca}}

%
\maketitle              
\begin{abstract}
When analyzing the spread of viruses, epidemiologists often need to identify the location of  infected hosts. This information can be found in public databases, such as GenBank~\cite{genebank}, however, information provided in these databases are usually limited to the country or state level. More fine-grained localization information requires phylogeographers to manually read relevant scientific articles. In this work we propose an approach to automate the process of place name identification from medical (epidemiology) articles. 
The focus of this paper is to propose a deep learning based model for toponym detection and experiment with the use of external linguistic features and domain specific information. 
The model was evaluated using a collection of $105$ epidemiology articles from PubMed Central~\cite{Weissenbacher2015} provided by the recent SemEval task $12$~\cite{semeval-2019-web}.
Our best detection model achieves an F1 score of $80.13\%$, a significant improvement compared to the state of the art of $69.84\%$.
These results underline the importance of  domain specific embedding as well as specific linguistic features in toponym detection in medical journals. 

\end{abstract}
\section{Introduction}
With the increase of global tourism and international trade of goods, phylogeographers, who study the geographic distribution of viruses, have observed an increase in the geographical spread of viruses~\cite{spread-of-vaccinated-diseases,trends-in-infectious-diseases}. In order to study and model the global impact of the spread of viruses, epidemiologists typically use information on the DNA sequence and structure of viruses, but also rely on meta data. Accurate geographical data is essential in this process. However, most publicly available data sets, such as GenBank \cite{genebank}, lack specific geographical details, providing information only at the country or state level. Hence, localized geographical information has to be extracted through a manual inspection of medical journals.

The task of toponym resolution is a sub-problem of named entity recognition (NER), a well studied topic in Natural Language Processing (NLP). Toponym resolution consists of two sub-tasks: toponym identification and toponym disambiguation. Toponym identification consists of identifying the word boundaries of expressions that denote geographic expressions; while toponym disambiguation focuses on labeling the expression with their corresponding geographical locations.  Toponym resolution has been the focus of much work in recent years (e.g.~\cite{ardanuy2017toponym,delozier2015gazetteer,taylor2017reduced}) and studies have shown that the task is highly dependent on the textual domain~\cite{web-geotagging-1,web-geotagging-2,web-geotagging-3,enc-geotagging,news-geotagging}. The focus of this paper is to propose a deep learning based model for toponym detection and experiment with the use of external linguistic features and domain specific information. The model was evaluated using the recent SemEval task $12$ datatset~\cite{semeval-2019-web} and shows that domain specific embedding as well as some linguistic features do help in toponym detection in medical journals.

\section{Previous Work}
\begin{figure}
\centering
\includegraphics[width=\linewidth]{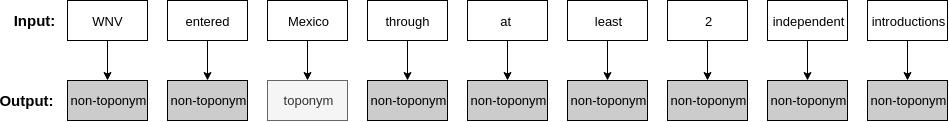}
\caption{\label{fig:input-output-example} An example of input and expected output of toponym detection task. Example from the~\cite{Weissenbacher2015}~dataset. }
\end{figure}

The task of toponym detection consists in labeling each word of a text as a toponym or non-toponym. For example, given the sentence:
\enumsentence{WNV entered Mexico through at least 2 independent introductions\footnote{Example from the~\cite{Weissenbacher2015}~dataset.}.} 
The expected output is shown in Figure~\ref{fig:input-output-example}.

Toponym detection has been addressed using a variety of methods: rule based approaches (e.g.~\cite{tamames2010}), dictionary or gazetteer-driven (e.g.~~\cite{lieberman2011multifaceted}), as well as machine learning approaches (e.g.~\cite{santos2015using}). Rule based techniques try to manually capture textual clues or structures which could indicate the presence of a toponym. However, these handwritten rules are often not able to cover all possible cases, hence leading to a relatively large number of false negatives. Gazetteer driven approaches (e.g.~\cite{lieberman2011multifaceted}), suffer from a large number of false positive identifications, since they cannot disambiguate entities that refer to geographical locations from other categories of named entities. For example in the sentence, \enumsentence{ Washington was unanimously elected President by the Electoral College in the first two national elections.} the word {\it Washington} will be recognized as a toponym since it is present in geographic gazetteers but in this context, the expression refers to a person name. 
Finally, standard machine learning approaches (e.g.~\cite{santos2015using}), require large datasets of labeled texts and carefully engineered features. Collecting such large datasets is costly and feature engineering is a time consuming task, with no guarantee that all relevant features have been modeled. This motivated us to experiment with automatic feature learning to address the problem of toponym detection. Deep Learning approaches to NER (e.g.~\cite{chiu2016named,collobert2008unified,lample2016neural,li2015biomedical,wang2015unified}) have shown how a system can infer relevant features and lead to competitive performances in that domain.
The task of toponym resolution for the epidemiology domain is currently the object of the SemEval $2019$ shared task $12$~\cite{semeval-2019-web}.
Previous approaches to toponym detection in this domain includes rule based approach~\cite{Weissenbacher2015}, Conditional Random Fields~\cite{weissenbacher2017}, and a mixture of deep learning and rule based approaches   ~\cite{detect}.
The baseline model used at the SemEval $2019$ task $12$~\cite{semeval-2019-web} is modeled after the deep feed forward neural network (DFFNN) architecture presented in~\cite{detect}. The network consists of $2$ hidden layers with $150$ rectified linear unit (ReLU) activation functions per layer. The baseline F1 performance is reported to be $69.84\%$. Building upon the work of
~\cite{semeval-2019-web,detect} we propose a DFFNN that uses domain-specific information as well as linguistic features to enhance the state of the art performance.

\section{Our Proposed Model}
\begin{figure}
\centering
\includegraphics[width=\linewidth]{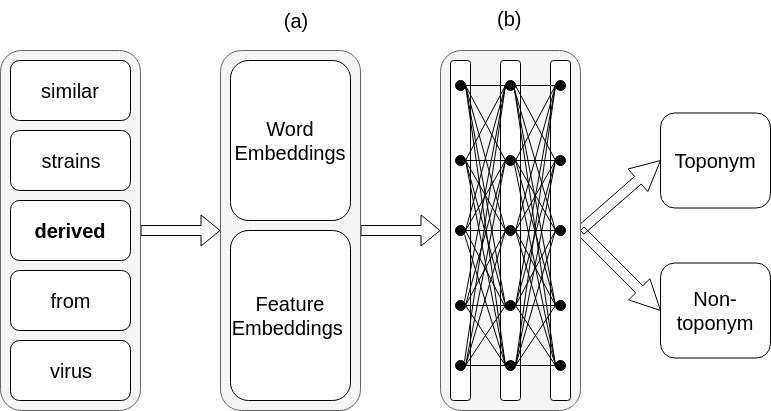}
\caption{\label{fig:architecture} Toponym recognition model. Input: words are extracted with a fixed context window (a) Embeddings: For each window, an embedding is constructed (b) Deep Neural Network: A feed-forward neural network with $3$ layers and $500$ neurons per layer outputs a prediction label indicating whether the word in the center of the context window is a toponym or not. }
\end{figure}

Figure~\ref{fig:architecture} shows the architecture of our toponym recognition model. The model is comprised of $2$ main layers: an embedding layer, and a deep feed-forward network.
\subsection{Embedding Layer \label{section:embedding-layer}}
As shown in Figure~\ref{fig:architecture}, the model takes as input a word (e.g. {\it derived}) and its context (i.e. $n$ words around it). Given a document, each word is converted into an embedding along with its context. Specifically, two types of embeddings are used: word embeddings and feature embeddings.

For word embeddings, our basic model uses the pretrained Wikipedia-PubMed embeddings\footnote{\url{http://bio.nlplab.org/}}. This embedding model was trained on a vocabulary of $201,380$ words and each word is represented by a $200$ dimensional feature vector. This embedding model was used as opposed to more generic Word2vec~\cite{word2vec} or GloVe~\cite{GloVe} in order to capture more domain specific information (see Section~\ref{section:experiment-resault}). Indeed, the corpus used for training the Wikipedia-PubMed embedding consists of Wikipedia pages and PubMed articles~\cite{word-emb}. This entails that the embeddings should be more appropriate when processing medical journals, and domain specific words. Moreover, the embedding model can better represent the closeness and relation of words in medical articles.
The word embeddings for the target word and its context are concatenated to form a single word embedding vector of size $200\times(2c+1)$, where $c$ is the context size.

Specific linguistic features have been shown to be very useful in toponym detection~\cite{detect}. In order to leverage this information, our model is augmented using embedding for these features. These include the use of capital letters for the first character of the word or for all characters of the word. These features are encoded as a binary vector representation. If a word starts with a capitalized letter, its feature embedding is $[1,0]$ otherwise it is $[0,1]$ and if all of its letters are capitalized then its feature embedding is $[1,1]$. 
Other linguistic features we observed to be useful (see  Section~\ref{section:experiment-resault}) include part of speech tags, and the word embedding of the lemma of the word. The feature embedding of the input word and its context are combined to the word embedding via concatenation to form a single vector and passed to the next layer.

\subsection{Deep Feed Forward Neural Network \label{sec:the-model}}

The concatenated embeddings formed in the embedding layer (Section~\ref{section:embedding-layer}) are fed to a deep feed forward network (DFFNN) (see Figure~\ref{fig:architecture}) whose task is to perform binary classification. This component is comprised of $3$ hidden layers and one output layer. Each hidden layer is comprised of $500$ ReLU activation nodes. Once an input vector $x$ enters a hidden layer $h$, the output $h(x)$ is computed as:
\begin{equation}\label{eq:hidden-layer}
    h(x) = \mathrm{ReLU}(Wx+b)
\end{equation}
The model is defined using the above equation recursively for all $3$ hidden layers. The output layer contains $2$ dimensional softmax activation functions. Upon receiving the input $x$, this layer will output $O(x)$ as follows:
\begin{equation}
    \label{eq:softmax}
    O(x)=\mathrm{Softmax}(Wx+b)
\end{equation}
The Softmax function was chosen for the output layer since it provides a categorical probability distribution over the labels for an input $x$, i.e.:
\begin{equation}
    p(x=\text{toponym})=1-p(x=\text{non-toponym})
\end{equation}
We employed $2$ mechanisms to prevent overfitting: drop-out and early-stopping. In each hidden layer the probability of drop-out was set to $0.5$. The early-stopping caused the training to stop if the loss on the development set (see Section~\ref{section:experiment-resault}) started to rise preventing over-fitting and poor generalization. Norm clipping~\cite{pascanu2013difficulty} scales the gradient when its norm exceeds a certain threshold and prevents the occurrence of exploding gradient; we experimentally found the best performing threshold to be $1$ for our model. 

We experimented with variations of the model architecture both in depth and number of hidden units per layer as well as other hyper-parameters listed in Table~\ref{table:hyper-parameters}. However, deepening the model lead to immediate over-fitting due to the small size of the dataset used~\cite{over-fitting} (see Section~\ref{section:experiment-resault}) even with the presence of a dropout function to prevent it. The optimal hyper-parameter configuration with the development set used to fine tune them can be found in Table~\ref{table:hyper-parameters}.

\begin{table}
\caption{\label{table:hyper-parameters} Optimal hyper-parameters of the neural network.}
\small
\centering
\begin{tabular}{l c}
\hline 
Parameters & Value\\
\hline 
Learning Rate & $0.01$\\
Batch Size & $32$\\
Optimizer & SGD\\
Momentum & $0.1$\\
Loss & Weighted Categorical cross-entropy\\
Loss weights & $(2,1)$ for toponym vs. nontoponym\\
\hline
\end{tabular}
\end{table}

\section{Experiments and Results \label{section:experiment-resault}}
Our model has been evaluated as part of the recent SemEval 2019 task 12 shared task~\cite{semeval-2019-web}. As such, we used the dataset and the scorer\footnote{\url{https://competitions.codalab.org/competitions/19948\#learn_the_details-evaluation}} provided by the organisers. The dataset consists of $105$ articles from PubMed annotated with toponym mentions and their corresponding geographical locations. 
The dataset was split into 3 sections: training, development, and test set containing $60\%$, $10\%$, and $30\%$ of the dataset respectively. Table~\ref{table:data-stats} shows statistics of the dataset.

\begin{table}
\caption{\label{table:data-stats} Statistics of the dataset.}
\small
\centering
\begin{tabular}{l r | r | r }
\hline 
& \textbf{Training} & \textbf{Development} & \textbf{Test}\\
\hline
Size &$2.8$MB&$0.5$MB&$1.5$MB\\
Number of articles & $63$ & $10$  & $32$\\
Average size of each article (in words)& $6422$ & $5191$ & $6146$\\
Average number of toponyms per article & $43$ & $44$ & $50$\\
\hline
\end{tabular}
\end{table}

A baseline model for toponym detection was also provided by the organizers for comparative purposes. The baseline, inspired by \cite{detect}, also uses a DFFNN but only uses $2$ hidden layers and $150$ ReLU activation functions per layer. 

Table~\ref{table:detect-results} shows the results of our basic model presented in Section~\ref{sec:the-model} (see $\# 4$) compared to the baseline (row $\#3$).\footnote{At the time of writing this paper, the results of the other teams were not available. Hence only a comparison with the baseline can be made at this point.}We carried out a series of experiments to evaluate a variety of parameters. These are described in the next sections. 
\begin{table}
\caption{\label{table:detect-results} Performance score of the baseline, our proposed model and its variations. The suffixes represent the presence of a feature, P.:Punctuation marks, S: Stop words, C: Capitalization features, POS: Part of speech tags, W: Weighted loss, L: Lemmatization feature. For example DFFNN Basic$+$P$+$S$+$C$+$POS refers to the model that only takes advantage of capitalization feature and part of speech tags and does not ignore stop words or punctuation marks.}
\small
\centering
\begin{tabular}{l l r r r r}
\hline 
\textbf{\#} & \textbf{Model} & \textbf{Context} & \textbf{Precision} & \textbf{Recall} & \textbf{F1}\\
\hline
$8$& DFFNN Basic$+$P$+$S$+$C$+$POS$+$W$+$L &5 & $80.69\%$ & $79.57\%$ & $80.13\%$\\
$7$& DFFNN Basic$+$P$+$S$+$C$+$POS$+$W &5 & $76.84\%$ & $77.36\%$ & $77.10\%$\\
$6$& DFFNN Basic$+$P$+$S$+$C$+$POS &5  & $77.55\%$ & $70.37\%$ & $73.79\%$\\
$5$& DFFNN Basic$+$P$+$S$+$C  &2 & $78.82\%$ & $66.69\%$ & $72.24\%$\\
$4$& DFFNN Basic$+$P$+$S &2 & $79.01\%$ & $63.25\%$ & $70.26\%$\\
$3$& Baseline &2 & $73.86\%$ & $66.24\%$ & $69.84\%$\\
$2$& DFFNN Basic$+$P$-$S  &2 & $74.70\%$ & $63.57\%$ & $68.67\%$\\
$1$& DFFNN Basic$+$S$-$P &2 & $64.58\%$ & $64.47\%$ & $64.53\%$\\
\hline
\end{tabular}
\end{table}
\subsection{Effect of Domain Specific Embeddings}

As~\cite{web-geotagging-1,web-geotagging-2,web-geotagging-3,enc-geotagging,news-geotagging} showed, the task of toponym detection is dependent on the discourse domain; this is why our basic model used the Wikipedia-PubMed embeddings. In order to measure the effect of such domain specific information, we experimented with $2$ other pretrained word embedding models: Google News Word2vec~\cite{Google-Word2vec}, and a GloVe Model trained on Common Crawl~\cite{GloVe-web}. Table~\ref{table:word-emb-details} shows the characteristics of these pretrained embeddings. Although, the Wikipedia-PubMed has a smaller vocabulary in comparison to the other embedding models, it suffers from the smallest percentage of out of vocabulary (OOV) words within our dataset since it was trained on a closer domain.
\begin{table}
\caption{\label{table:word-emb-details} Specifications of the word embedding models. }
\small
\centering
\begin{tabular}{l r | c | c}
\hline 
\textbf{Model} & \textbf{Vocabulary Size} & \textbf{Embedding Dimension} & \textbf{OOV words}\\
\hline
Wikipedia-PubMed & $201,380$ & $200$ & $28.61\%$\\
Common Crawl GloVe & $2.2$M & $300$ & $29.84\%$\\
Google News Word2vec & $3$M & $300$ & $44.36\%$\\
\hline
\end{tabular}
\end{table}

We experimented with our DFFNN model with each of these embeddings and optimized the context window size to achieve the highest F-measure on the development set. The performance of these models on the test set is shown in Table~\ref{table:word-emb-performance}. As predicted, we observe that Wikipedia-PubMed performs better than the other embedding models. This is likely due to its small number of OOV words and its domain-specific knowledge. As Table~\ref{table:word-emb-performance} shows, the performance of the GloVe model is quite close to the performance of Wikipedia-PubMed. To investigate this further, we decided to combine the two embeddings and train another model and evaluate performance. As shown in Table~\ref{table:word-emb-performance}, the performance of this model (Wikipedia-PubMed $+$ GloVe) is higher than the GloVe model alone but lower than the Wikipedia-PubMed. This decrease in performance suggests that because GloVe embeddings are more general, when the network is presented with a combination of GloVe and Wikipedia-PubMed, they dilute the domain specific information captured by the Wikipedia-PubMed embeddings, hence the performance suffers.  
From here on, our experiments were carried on using Wikipedia-PubMed word embeddings alone.
\begin{table}
\caption{\label{table:word-emb-performance} Effect of word embeddings on the performance of our proposed model architecture.}
\small
\centering
\begin{tabular}{l c c c c}
\hline 
\textbf{Model} & \textbf{Context Window} & \textbf{Precision} & \textbf{Recall} & \textbf{F1}\\
\hline
Wikipedia-PubMed & $2$ & $79.01\%$ & $63.25\%$ & $70.26\%$\\
Wikipedia-PubMed $+$ GloVe & $2$ & $73.09\%$ & $67.22\%$ & $70.03\%$\\
Common Crawl GloVe & $1$ & $75.40\%$ & $64.05\%$ & $69.25\%$\\
Google News Word2vec & $3$ & $75.14\%$ & $58.96\%$ & $66.07\%$\\
\hline
\end{tabular}
\end{table}

\subsection{Effect of Linguistic Features}
Although deep learning approaches have lead to significant improvements in many NLP tasks, simple linguistic features are often very useful. In the case of NER, punctuation marks constitute strong signals. To evaluate this in our task, we ran the DFFNN Basic without punctuation information. As Table~\ref{table:detect-results} shows, the removal of punctuation, decreased the F-measure from $70.26\%$ to $64.53\%$ (see Table~\ref{table:detect-results} $\# 1$). A manual error analysis showed that many toponyms appear inside parenthesis, near a dot at the end of a sentence, or after a comma. Hence, as shown in~\cite{gelernter2013algorithm} punctuation is a good indicator of toponyms and should not be ignored.

As Table~\ref{table:detect-results} ($\# 2$) shows,  the removal of stop words, did not help the model either and lead to a decrease in F-measure (from $70.26\%$ to $68.67\%$). We hypothesize that some stop words such as \textit{in} do help the system detect toponyms as they provide a learnable structure for detection of toponyms and that is why the model accuracy suffered once the stop words were removed. 

As seen in Table~\ref{table:detect-results} our basic model suffers from low recall. A manual inspection of the toponyms in the dataset revealed that either their first letter is capitalized (e.g. \textit{Mexico}) or all their letters are capitalized (e.g. \textit{UK}). As mentioned in Section~\ref{section:embedding-layer} we used this information in an attempt to help the DFFNN learn more structure from the small dataset. As a result the F1 performance of the model increased form $70.26\%$ to $72.27\%$ (see Table~\ref{table:detect-results} $\# 5$).

In order to help the neural network better understand and model the structure of the sentences, we experimented with part of speech (POS) tags as part of our feature embeddings. We used the NLTK POS tagger~\cite{bird2009natural} which uses the Penn Treebank tagset. As shown in Table~\ref{table:detect-results} ($\# 6$), the POS tags significantly improve the recall of the network (from $66.69\%$ to $70.37\%$) hence leading to a higher performance in F1 (from $72.24\%$ to $73.79\%$).
The POS tags help the DFFNN to better learn the structure of the sentences and take advantage of more contextual information (see Section~\ref{sec:window}).

\subsection{\label{sec:window} Effect of Window Size}
\begin{figure}
\centering
\includegraphics[width=\linewidth, height=7cm]{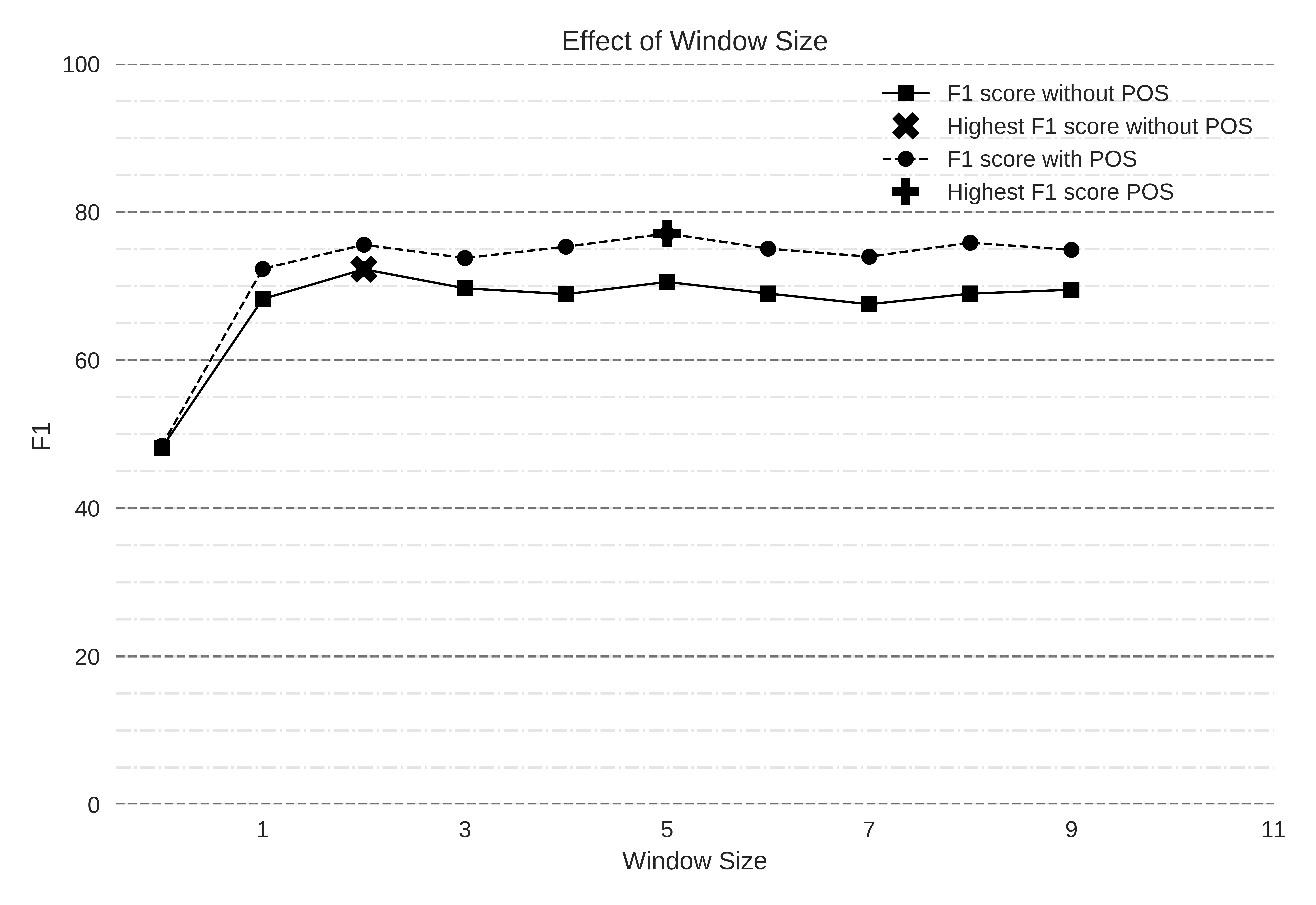}
\caption{\label{fig:widnow-exp} Effect of context window on performance of the model with and without POS features. (DFFNN Basic$+$P$+$S and DFFNN Basic$+$P$+$S$+$C$+$P )}
\end{figure}

In order to measure the effect of the size of the context window, we varied this value using the basic DFFNN. As seen in Figure~\ref{fig:widnow-exp}, the best performance is achieved at $c=2$. With values over this threshold, the DFFNN overfits as it cannot extract any meaningful structure. Due to the small size of the data set, the DFFNN is not able to learn the structure of the sentences, hence increasing the context window alone does not help the performance. In order to help the neural network better understand and use the contextual structure of the sentences in its predictions, we experimented with part of speech (POS) tags as part of our feature embeddings. As shown in Figure~\ref{fig:widnow-exp}, the POS tags help the DFFNN to take advantage of more contextual information as a result the DFFNN with POS embeddings achieves a higher performance on larger window sizes. The context window for which the DFFNN achieved its highest performance on the development set was $c=5$, and on the test set the performance was increased from $72.24\%$ to $77.10\%$ (see Table~\ref{table:detect-results} $\# 6$).

\subsection{Effect of the Loss Function}

As shown in Table~\ref{table:data-stats} most models suffer from a lower recall than precision. The dataset is quite imbalanced, that is the number of non-toponyms are much higher than toponyms (99\% vs 1\%). Hence, the neural network prefers to optimize its performance by concentrating its efforts on correctly predicting the labels for the dominant class (non-toponym).  In order to minimize the gap between recall than precision, we experimented with a weighted loss function. We adjusted the importance of predicting the correct labels experimentally and found that by weighing the toponyms 2 times more than the non-toponyms, the system reaches an equilibrium in the precision and recall measure, leading to a higher F1 performance. (This is indicated by \say{w} in Table~\ref{table:detect-results} row $\# 7$) 

\subsection{Use of Lemmas}

Neural networks require large datasets to learn structures and they learn better if the dataset contains similar examples so that the system can cluster them in its learning process. Since our dataset is small and the Wikipedia-PubMed embeddings suffer from $28.61\%$ OOV words (see Table~\ref{table:word-emb-details}), we tried to help the network better cluster the data by adding the lemmatized word embeddings of the words to the feature embeddings and see how our best model reacts to it. As shown in Table~\ref{table:detect-results} ($\# 8$), this improved the F1 measure significantly (from $77.10\%$ to $80.13\%$).

\begin{figure}
\centering
\includegraphics[width=\linewidth]{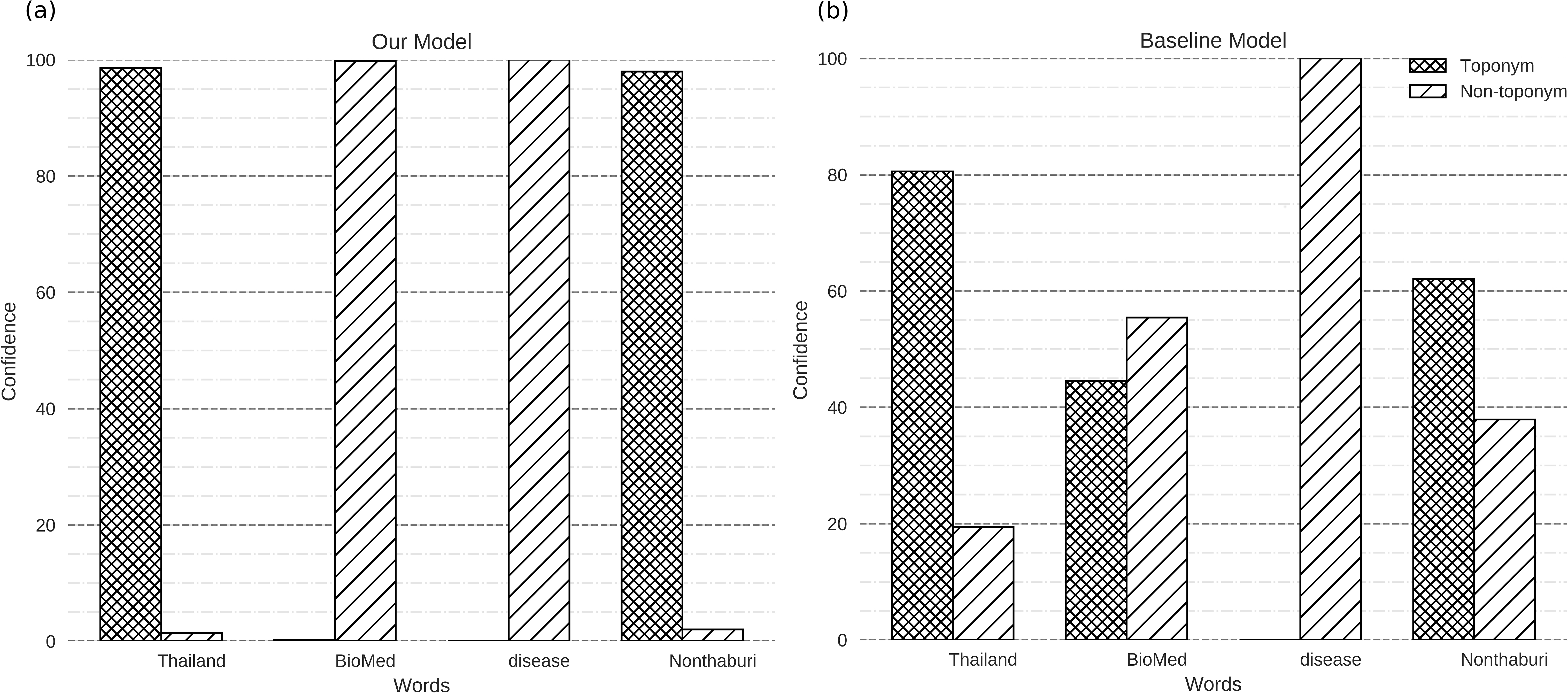}
\caption{\label{fig:pred-confidence} (a) Confidence of our proposed model in its categorical predictions. (b) Confidence of the baseline in its categorical predictions.}
\end{figure}

Furthermore, we picked $2$ random toponyms and $2$ random non-toponyms to visualize the confidence of our best model and the baseline model in their prediction as given by the softmax function (see Equation~\ref{eq:softmax}). Figure~\ref{fig:pred-confidence} shows that our model produces much sharper confidence in comparison to the baseline model.\\

\section{Discussion}
Overall our best model (DFFNN $\# 8$ in Table~\ref{table:detect-results}) is made out of the basic DFFNN plus capitalized feature, POS embeddings, weighted loss function, and lemmatization feature.
The experiments and results described in Section~\ref{section:experiment-resault} underlines the importance of linguistic insights in the task of toponym detection. Ideally the system should learn all these insights and features by itself given access to enough data. However, when the data is scarce, as in our case, we should take advantage of the linguistic structure of the data for better performance.

Our experiments also underline the importance of domain specific word embedding models. These models reduce OOV words and also present us with embeddings that capture the relation of the words in the specific domain of study.
\section{Conclusion and Future Work}
This paper presented the approach we used to participate to the recent SemEval task 12 shared task on toponym resolution~\cite{semeval-2019-web}. Our best DFFNN approach took advantage of domain specific embeddings as well as linguistic features. It achieves a significant increase in F-measure compared to the baseline system (from $69.74\%$ to $80.13\%$). However, as the official results were not available at the time of writing, comparison with other approaches cannot be done at this time.

The focus of this paper was to propose a deep learning based model for toponym detection and experiment with the use of external linguistic features and domain specific information. The model was evaluated using the recent SemEval task $12$ datatset~\cite{semeval-2019-web} and shows that domain specific embedding as well as some linguistic features do help in toponym detection in medical journals.

One of the main factors preventing us from exploring deeper models, was the small size of the data set. With more human annotated data the models could be extended for better performance. However, since human annotated data is expensive to produce, we suggest distant supervision~\cite{krause2012large} to be explored for further increasing performance. As our experiments pointed out, the model could heavily benefit from linguistic insights, hence equipping the model with more linguistic driven features could potentially lead to a higher performing model. We did not have the time or computational resources to explore recurrent neural architectures, however future work could be done focusing on these models.

\section*{Acknowledgments}

This work was financially supported by the Natural Sciences and Engineering Research Council of Canada (NSERC).

\bibliographystyle{splncs04}
\bibliography{paper}
\end{document}